\documentclass[a4paper,twocolumn]{article}
\usepackage{graphicx}

\title{Novelty Detection for Robot Neotaxis \footnote{In Proceedings of the 2nd International Symposium on Neural Computation, pages 554 - 559, 2000}}
\author{Stephen Marsland, Ulrich Nehmzow and Jonathan Shapiro\\
Department of Computer Science\\University of Manchester\\Oxford Road\\Manchester M13 9PL\\
\texttt{\{smarsland, ulrich, jls\}@cs.man.ac.uk}}

\date{}

\begin{document}
\maketitle

\thispagestyle{empty}
\begin{abstract}
The ability of a robot to detect and respond to
changes in its environment
is potentially very useful, as it draws
attention to new and potentially
important features. We describe an algorithm for
learning to filter out previously experienced
stimuli to allow further concentration
on novel features. The algorithm uses
a model of habituation, a biological process which
causes a decrement in response with repeated presentation.
Experiments with a 
mobile robot are presented in which the robot
detects the most novel stimulus and turns
towards it (`neotaxis').
\end{abstract}

\section{Introduction}

Many animals have the ability to detect novelty, that is 
to recognise new features or changes 
within their environment.
This paper describes an 
algorithm which learns to 
ignore stimuli which are presented repeatedly,
so that novel stimuli stand out.
A simple demonstration of the algorithm
on an autonomous mobile robot is given. 
We term the robot's behaviour of following the
most novel stimulus
{\it neotaxis}, meaning `turn towards
new things', taken from the Greek ({\it neo} = new,
{\it taxis} = follow). A number of different versions
of the novelty filter are described and compared to 
find the best for the particular data used. 

Attending to more novel
stimuli is a useful ability for a mobile robot as
it can limit the amount of data which the robot has to
process in order to deal with its environment. It can 
be used to recognise when perceptions are
new and must therefore be learned. In
addition, it means that the robot can be used as an
inspection agent, so that after training to learn
common features it will highlight any `novel' stimuli,
i.e., those which it has not seen previously.

\subsection{Related Work}

A number of novelty detection methods have been
proposed within the neural network literature,
but they are mostly trained off-line. Particularly
noteworthy is the Kohonen Novelty Filter~\cite{Kohonen76,Kohonen93},
which is an auto-encoder neural network trained by
back-propagation of error. After training, any presentation
to the network produces one of the trained outputs,
and the bitwise difference between the input and output
shows the novel parts of the input. This work has been
extended by a number of authors. For example, Aeyels~\cite{Aeyels90} adds 
a `forgetting' term into the equations.

Ho and Rouat~\cite{Ho98} use a biologically inspired model 
that times how long an oscillatory network takes to converge to a
stable output, reasoning that previously seen inputs should
converge faster than novel ones. Finally, Levine and Prueitt~\cite{Levine92}
use the gated dipole proposed by Grossberg~\cite{Grossberg72,Grossberg72a}
to compare inputs with pre-defined ones, novel features causing
greater output values. 

\section{The Novelty Filter}

\subsection{Habituation}

Habituation is a reduction in behavioural response that occurs
when a stimulus is presented to an organism repeatedly. It is 
present in many animals, from the sea slug {\it Aplysia}
~\cite{Bailey83,Greenberg87} through toads~\cite{Ewert78,Wang92} and cats~\cite{Thompson86}
to humans~\cite{OKeefe77}. It has been modelled by
Groves~\cite{Groves70}, Stanley~\cite{Stanley76}
and Wang and Hsu~\cite{Wang90}.
Habitation differs from other processes which
decrement synaptic efficacy, such as fatigue, 
in that a change in stimulus restores the response
to its original levels. This process is called
dishabituation. There is also a `forgetting'
effect, where a stimulus which has not been presented
for a long time recovers its response. Further
details can be found in~\cite{Thompson66,Wang95}.

\begin{figure}[h]
\centering
\includegraphics[angle=270,width=.45\textwidth]{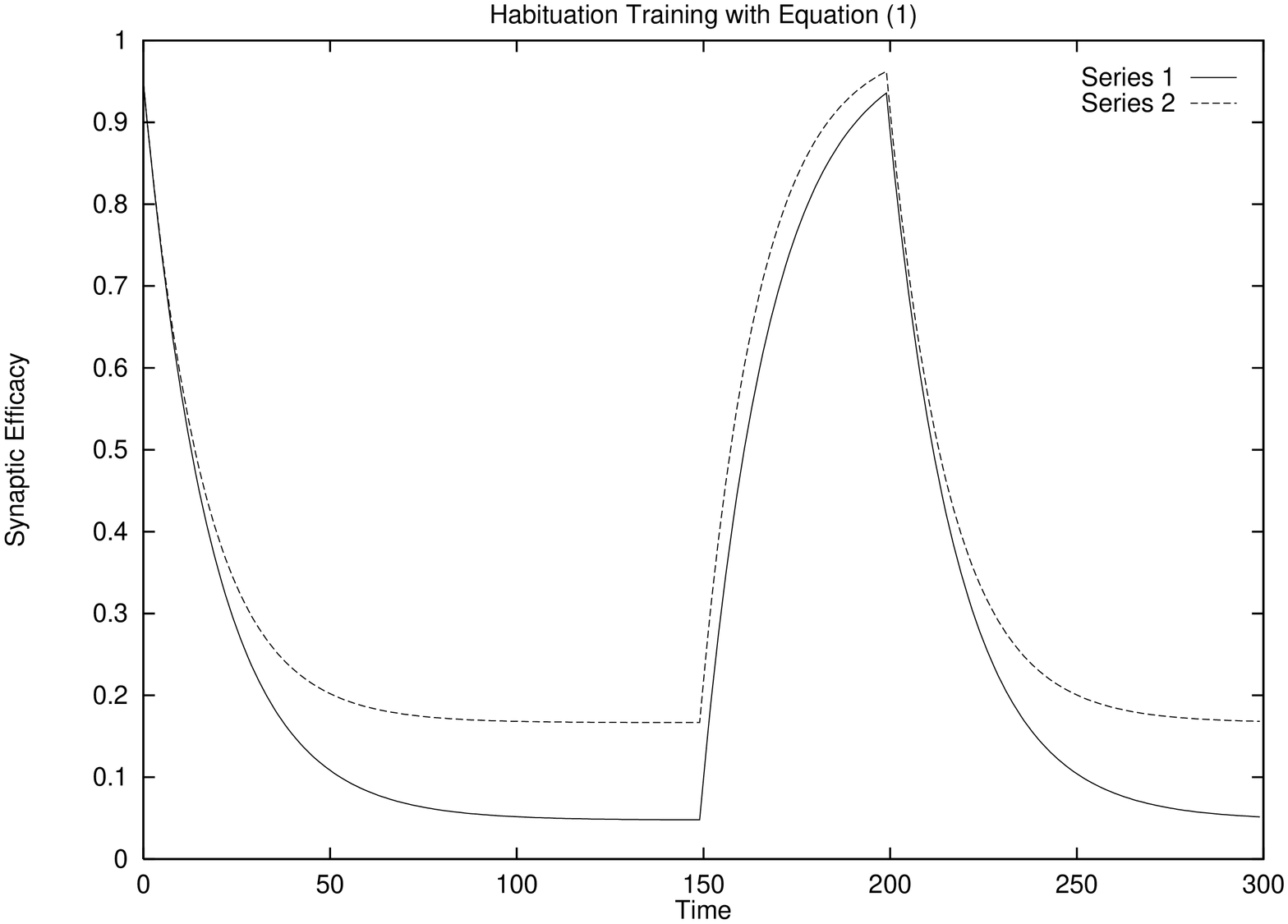}
\caption{
\textsf {\small An example of how the synaptic efficacy drops when habituation occurs. 
In the first, descending part of the graph, 
a stimulus $S(t)=1$ is presented continuously. This 
changes to $S(t)=0$ at $t=150$ where the synaptic efficacy 
rises again, and becomes $S(t)=1$ again at $t=200$, causing another drop.
The two curves show different values of the constants, in 
series 1 $\alpha = 1.05$ and in series 2 $\alpha = 1.2$. In both, 
$\tau = 20$ and $y_0 = 1.0$.}} 
\label{curves}
\end{figure}

The habituation mechanism used in the system
described here is Stanley's model.  The synaptic
efficacy, $y(t)$, decreases according to the
following equation:

\begin{equation}
 \tau \frac{dy(t)}{dt} = \alpha \left[ y_0 - y(t) \right] - S(t),
\label{HabEqn}
\end{equation}

\noindent
where $y_0$ is the original value of $y$, $\tau$ and $\alpha$
are time constants governing the rate of habituation and 
recovery respectively, and $S$ is the stimulus presented.
The effects of the equation are shown in figure~\ref{curves}.
The principal difference between this and the model of
Wang and Hsu is that the latter allows for long-term memory,
so repeated training causes faster learning.

Figure~\ref{curves} shows the synaptic efficacy increasing 
again at time 150, when the stimulus is removed. This
is effectively a `forgetting' effect, and is caused by a
dishabituation mechanism which increases the strength of
synapses that do not fire. In the implementation
described here this effect can be removed. The experiments 
reported in section~\ref{Results} investigate effects
of the filter both with and without forgetting.

\subsection{Using Habituation for a Novelty Filter\label{NF}}

\begin{figure}
\centering
\includegraphics[angle=270,width=.45\textwidth]{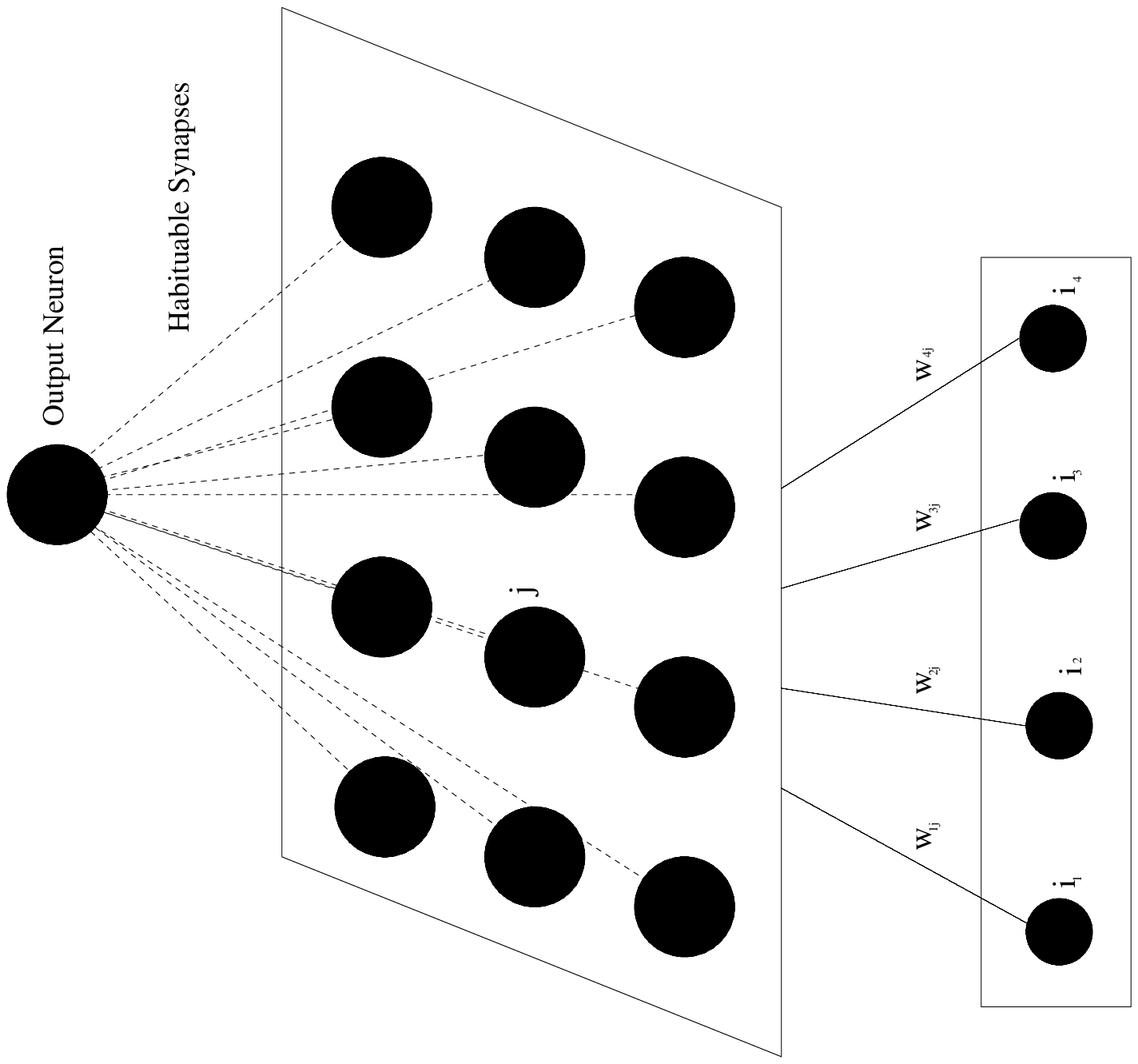}
\caption{ \textsf{ \small {The novelty filter.
The input layer connects to a clustering layer which
represents the feature space, the winning neuron (i.e.,
the one `closest' to the input) passing its
output along a habituable synapse to the output neuron so
that the output received from a neuron reduces with the
number of times it fires.}}}
\label{hsom}
\end{figure}

The principle behind the novelty filter is that perceptions
are classified by some form of clustering network,
whose output is modulated by
habituable synapses, so that the more frequently
a neuron fires, the lower the efficacy of the synapse
becomes. This means that only novel features will produce
any noticeable output. If the
habituable synapses receive zero input (rather than none) during turns
when their neuron does not fire, the synapses will
`forget' the inhibition over time, providing that 
this forgetting mechanism (or dishabituation) is turned on.

The choice of clustering algorithm is very important
and depends on the data being classified. In this
paper, we compare the performance of three different
networks, described below, on the robot application.
The three networks described were chosen because they
performed best on sample data that was selected to
be similar to that they would see on the robot. 
In addition to those described below,
the Neural Gas~\cite{Martinetz93}
network also performed well, but computational
constraints means that it was not possible to run
it on the robot.

\subsection{Some Possible Clustering Networks \label{NNs}}

\subsubsection{Kohonen's Self-Organising Map (SOM)}

Kohonen's Self-Organising Map~\cite{Kohonen93} works in the following way:\\
Every element of the input vector is connected to every
node of the map by a modifiable connection. The distance $d$
between the input and each of the neurons in the field
is calculated using
\begin{equation}
d = \sum_{i=0}^{N-1} \left[ \mathbf{v} (t) - \mathbf{w}_i (t) \right] ^2
\end{equation}

\noindent
where $\mathbf{v} (t)$ is the input vector at time $t$ and
$\mathbf{w}_{i}$ the weight between input $i$ and the neuron.
In a Learning Vector Quantiser~\cite{Kohonen93}, used here, the neuron with 
the minimum $d$ is selected and the
weight for that neuron and its topological neighbours
are updated by:

\begin{equation}
\mathbf{w}_{i} (t+1) = \mathbf{w}_{i} (t) + \eta (t) \left[ \mathbf{v} (t) - \mathbf{w}_{i} (t) \right]
\end{equation}

\noindent
where $\eta$ is the learning rate, $0 \leq \eta \leq 1$.

Usually, a two-dimensional SOM is used, but in the
implementation described here a ring-shaped network, effectively a line with the
end neurons linked together, was used.
The neighbourhood size and
learning rate remained constant so that the system was
always learning. The neighbourhood comprised
only the nearest neighbours of each neuron, and $\eta$
was fixed at 0.25.

\subsubsection{The Temporal Kohonen Map (TKM)}

This self-organising map, proposed by
Chappell and Taylor~\cite{Chappell93}, is 
based on Kohonen's SOM, but uses ``leaky integrator''
neurons whose activity decays exponentially over time. 
The exponential decay is controlled by a time
constant ($\gamma$ in equations~\ref{TKMeqn} and~\ref{TKMupdate} below). This is
similar to a short-term memory, allowing previous inputs
to have some effect on the processing of the current
input, so that the neurons which have won recently are
more likely to win again. In the experiments
reported here the value $\gamma = 0.4$ was used,
meaning that only the previous 2 or 3 winners had
any influence in deciding the current winner.
The activity of the neurons is calculated using 

\begin{equation}
a_i (t) = \gamma \cdot a_i (t-1) + e^{ \left( - \frac{1}{2} \right) \left[ \mathbf{v} (t) - \mathbf{w}_i (t) \right]^2},
\label{TKMeqn}
\end{equation}

\noindent and, in a similar way to the SOM, the neuron with the
largest activity $a$ is chosen as winner, and its weights and
those of its topological neighbours updated using the following 
weight update rule ($\eta$ and the neighbourhood remained the same):

\begin{equation}
\mathbf{w}_i (t+1) = \mathbf{w}_{i} (t) + \eta \sum_{k=0}^{n} \gamma^k \left[ \mathbf{v} (t-k) - \mathbf{w}_i (t-k) \right].
\label{TKMupdate}
\end{equation}

\subsubsection{The $K$--Means Clustering Algorithm}

One of the simplest ways to cluster data is by using the $K$--means
algorithm~\cite{Bishop95b}. A pre-determined number of prototypes, $\mathbf{\mu}$,
are chosen to represent the data, so that it is 
partitioned into $K$ clusters.
The positions of the prototypes are chosen to minimise the sum-of-squares
clustering function,

\begin{equation}
J = \sum_{j=1}^{K} \sum_{n \in S_j} \| \mathbf{x}^n - \mathbf{\mu}_j \|^2
\end{equation}
\noindent
for data points $\mathbf{x}^n$. This separates the data into $K$ partitions $S_j$.
The algorithm can be carried out as an on-line or batch procedure, with the on-line
version, used here, having the update rule

\begin{equation}
\Delta \mathbf{\mu}_j = \eta \left( \mathbf{x}^n - \mathbf{\mu}_j \right).
\end{equation}

\section{Using the Novelty Filter on a Mobile Robot\label{impl}}

The robot implementation was designed to show that the
novelty filter described in section~\ref{NF} can be used to detect new stimuli. 
The novelty filter was incorporated into a system
where a robot detects and turns towards new stimuli.
It was
implemented on a Fischer Technik mobile
robot, which uses a Motorola 68HC11 microcontroller.
The robot has a two wheel differential drive system and
four light sensors facing in the cardinal directions.

\begin{figure}
\centering
\includegraphics[width=.3\textwidth]{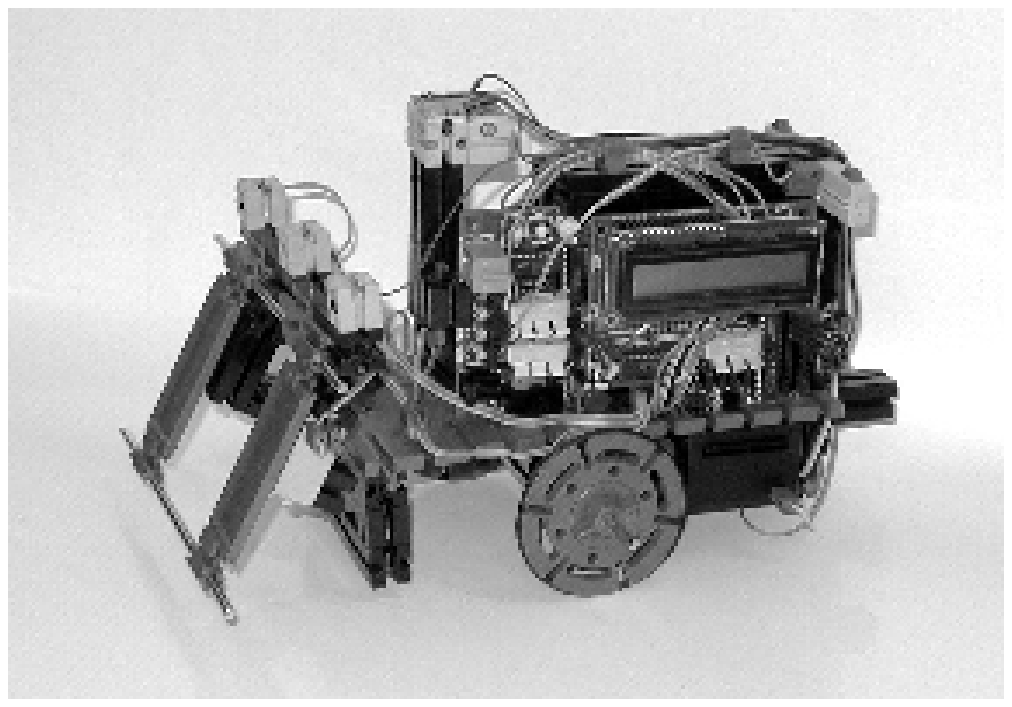}
\caption{ \textsf{ \small{The Fischer Technik Robot used in 
the experiments. The light sensors used in the experiments can be 
seen at the top of the mast towards the front of the robot.}}}
\end{figure}

\begin{figure*}[hbt]
\centering
\includegraphics[angle=270]{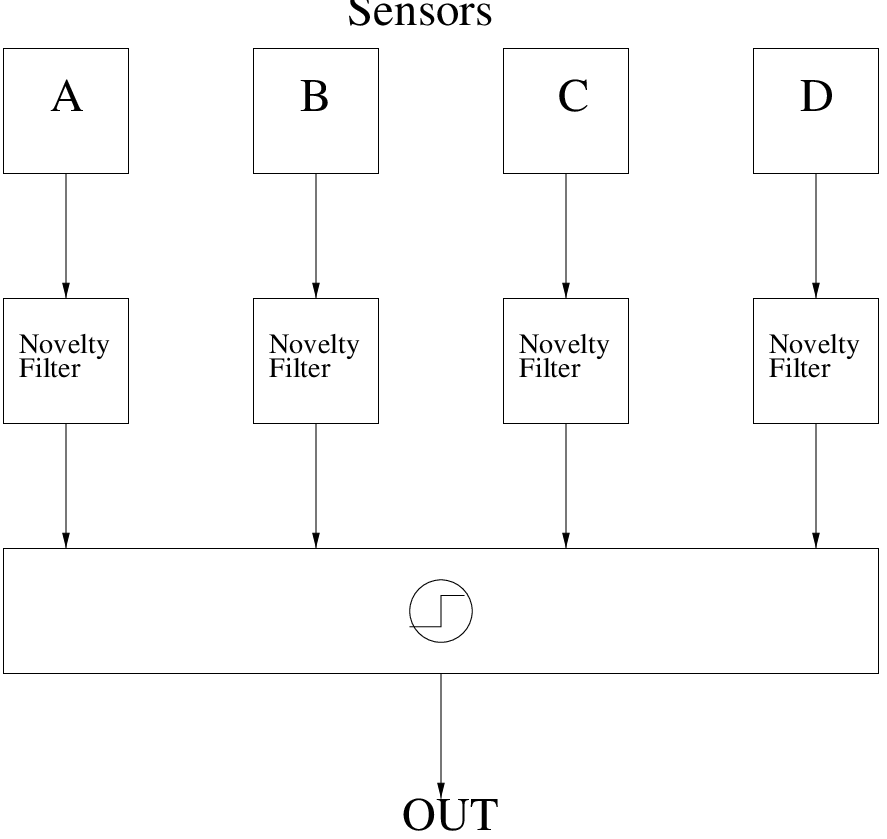}
\caption{ \textsf{ \small{The overall system for choosing the most interesting stimulus. 
Each sensory perception is classified separately by a novelty filter -- which receives
an input of present and recent perceptions -- and a value indicating
the novelty of that stimulus is output. Completely new
stimuli are given a higher priority. The most novel stimulus was selected
for a response, providing it exceeded a pre-defined threshold.}}}
\label{SysLayout}
\end{figure*}

In the experiments described below, the robot received a number
of different light stimuli, which varied in the frequency of the flashes.
It classified these stimuli autonomously
and decided whether or not to respond (turn towards the source)
according to how novel they were. 
Each of the sensors on the robot, in this case four light
sensors, had its own novelty filter,
as shown in figure~\ref{SysLayout}.
At each cycle, the current reading on each sensor was concatenated
with the previous five to form a six element input vector, known
as a delay line or lag vector. This
vector was classified by the novelty filter and an output produced. 
In the case of the TKM, which keeps an internal
history of previous inputs, only the most 
recent reading was needed as input. 

The output of the filter was a function of how many times that neuron had fired
before, due to the habituating synapse. 
Each of the four novelty filters fed their output
to a comparator function which propagated the strongest signal, 
providing that it was above a pre-defined threshold, to the
action mechanism.
If none of the stimuli
were strong enough, the cycle repeated. 
Owing to memory constraints, the clustering
mechanism was limited to just twelve neurons arranged 
in a ring. All three of the networks
described in section~\ref{NNs} were the same size.

A bypass function was associated with each
sensor. If a neuron had not fired before (that is, its
synapse had not been habituated) the comparator function
favoured it, so that the system responded
rapidly to new signals. If two new signals were
detected simultaneously, the stronger one was used.

\section{Experiments and Results \label{Results}}

\begin{figure}
\centering
%\hspace{10mm}
\includegraphics[angle=270,width=.2\textwidth]{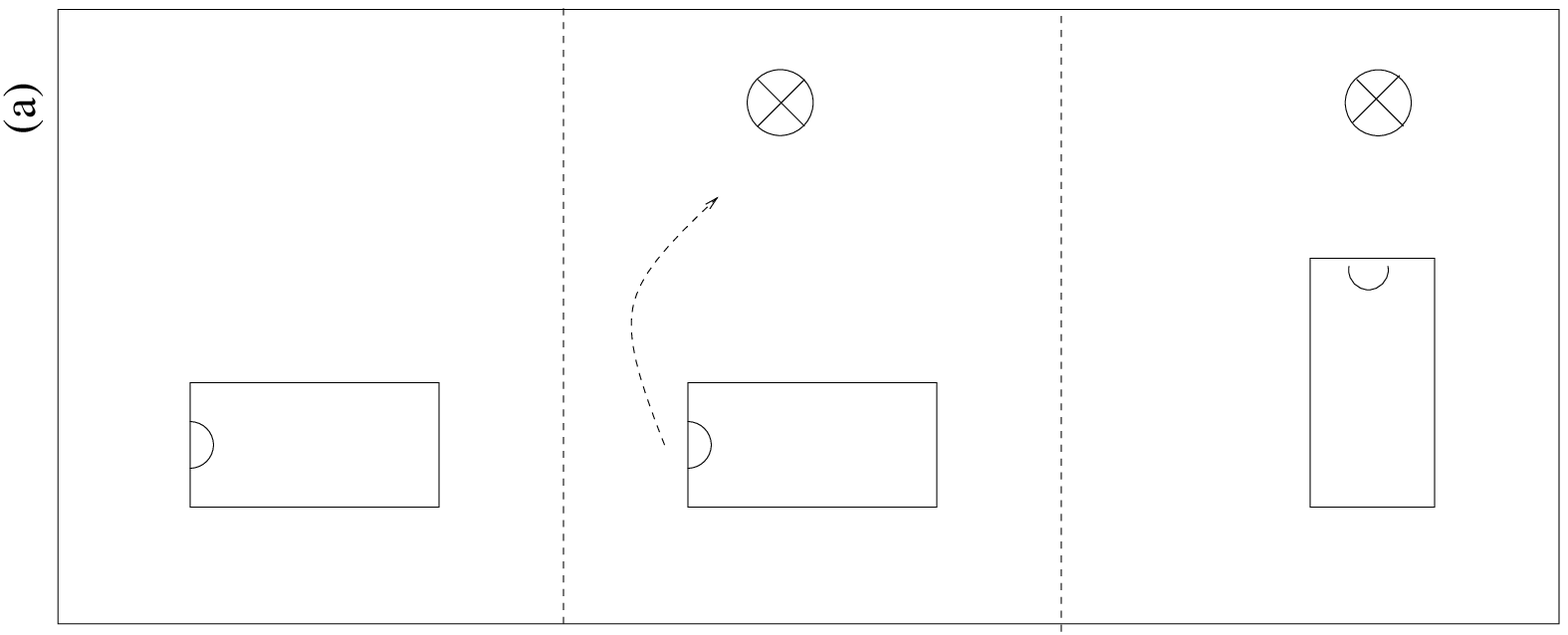}%
\hfill
\includegraphics[angle=270,width=.2\textwidth]{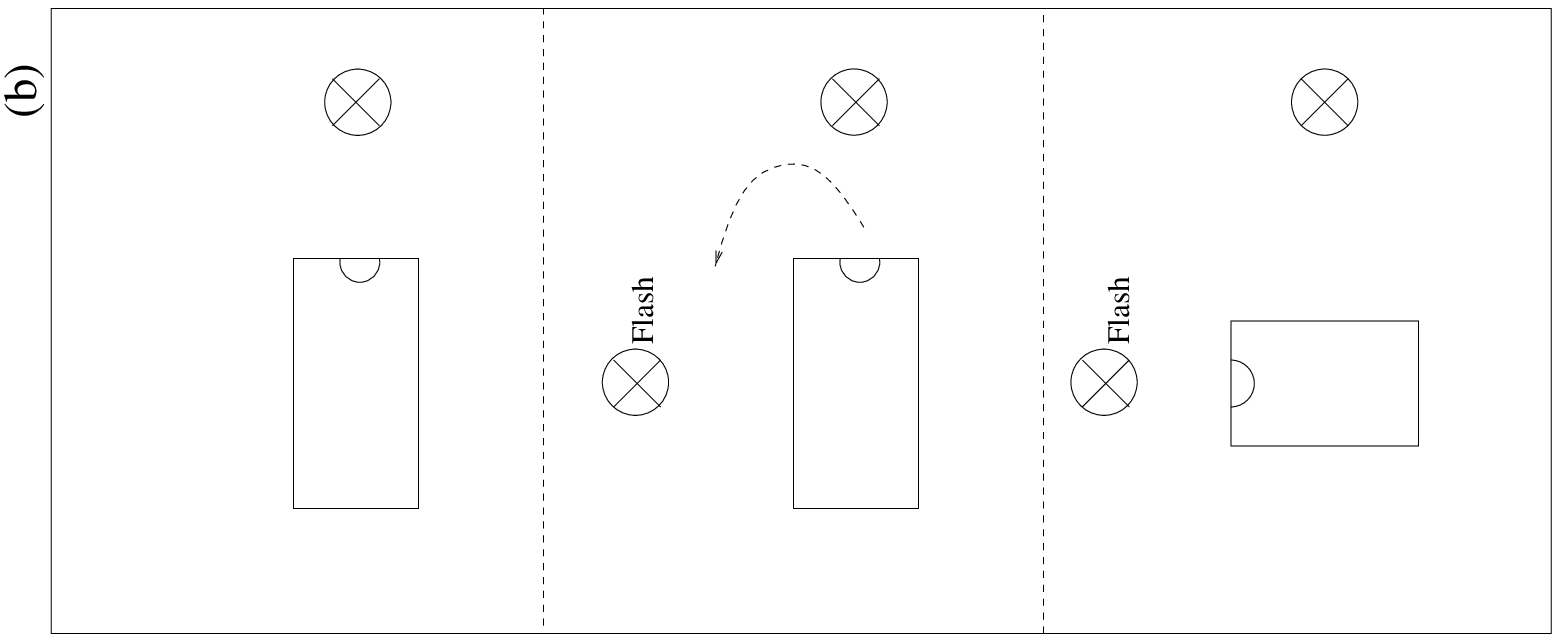}
\hspace{20mm}

\vspace{10mm}
%\hspace{10mm}
\includegraphics[angle=270,width=.2\textwidth]{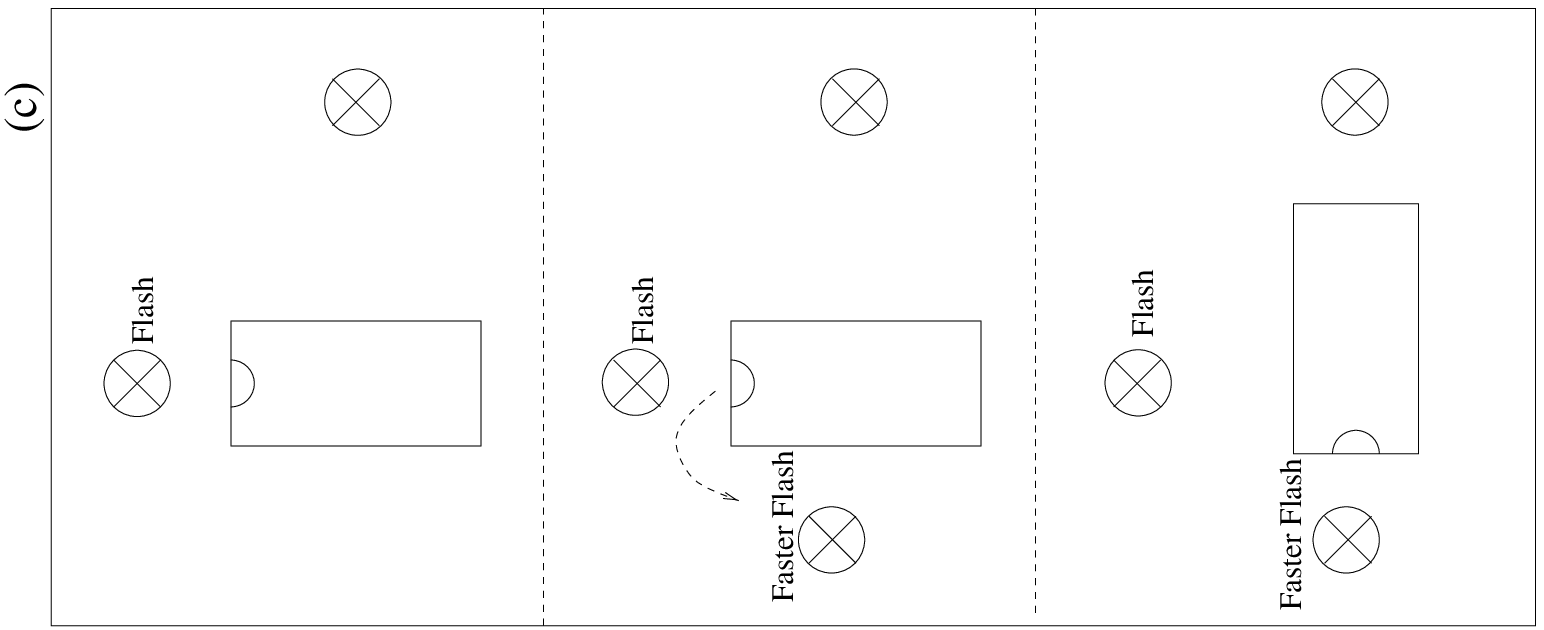}%
\hfill
\includegraphics[angle=270,width=.2\textwidth]{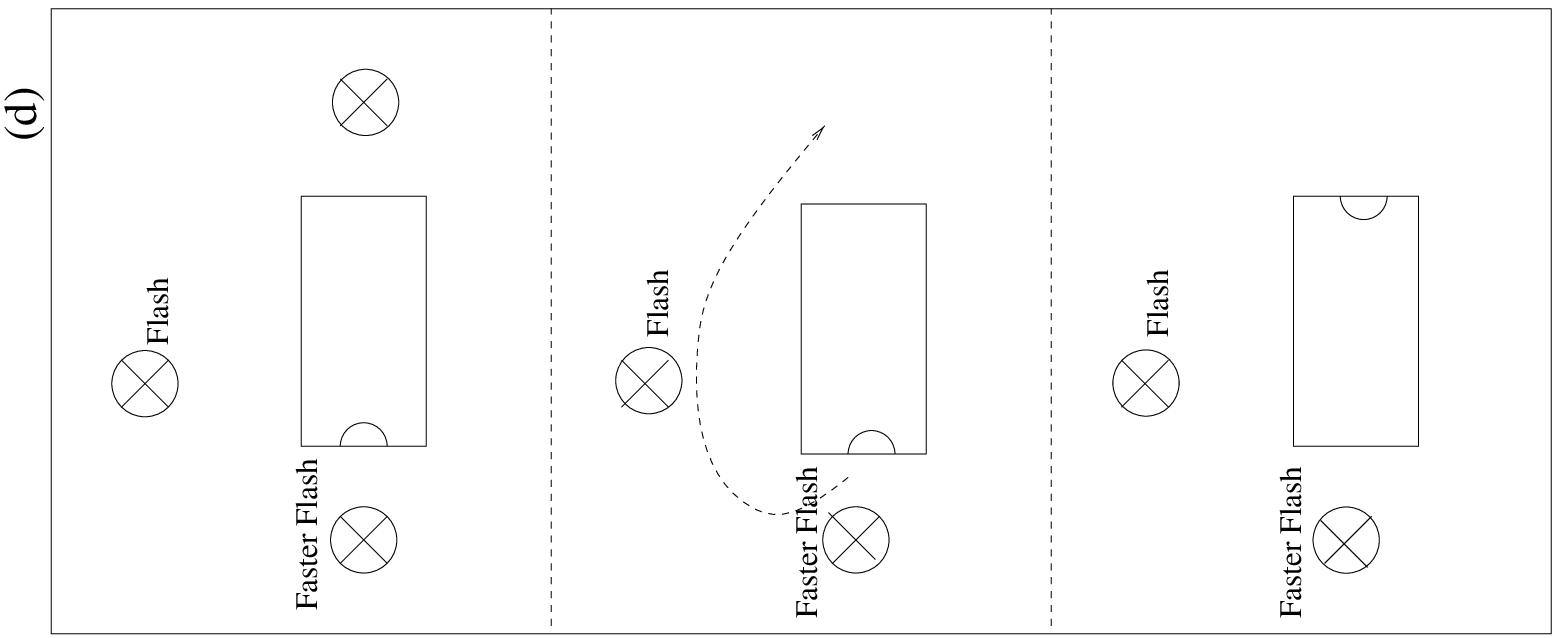}
\hspace{20mm}
\caption{ \textsf{ \small{ Figures showing the behaviour of the
robot during the four stages of the first experiment with forgetting.
The motion of the robot is shown using the dotted lines.
In (a) the robot turns towards the new light, in (b) it turns
towards the newer flashing light, and then in (c) to the
faster flashing light. Finally in (d) it turns back
to the point where the light has been turned off.}}}
\label{ExpLayout}
\end{figure}

Three separate experiments were carried out. The 
first, the results of which are shown in figure~\ref{ExpLayout} and table~\ref{Tab1}, was designed
to test the forgetting mechanism as well as the
general ability to turn towards novel stimuli. 
The robot was initially placed in a featureless environment. 
A light was 
introduced to project onto one of the light sensors.
Once the robot had turned to face this light
source, a second, slowly flashing light was added.
As this light was more novel, the robot 
turned towards it. A further, faster flashing light was
then introduced, which the robot again faced.
Finally, the
constant light was switched off and, 
in the case where a `forgetting' mechanism was used,
the robot perceived this lack of stimulus 
as novel and turned back towards it. Otherwise it
did not respond.

In the second experiment, steps (a) and (b) of
figure~\ref{ExpLayout} were again followed. However,
instead of a faster flash being shown in the
third stage, a second flashing light of the same
(slow) frequency was shown. If 
the flashing light was still novel, the robot
turned towards this as it was a newer version
of the most novel stimulus. However, if the 
flashing light had ceased to be novel, the robot
ignored it.

Finally, instead of a second flashing light in
part (c), a second constant light was introduced. 
Whether or not the robot responded to this 
depended on whether or not the forgetting mechanism
was switched on and which sensor it was on -- if it
was a sensor which had not previously seen it, the robot
responded.

Table~\ref{Tab1} shows the reactions of the robot
in the three experiments, both with forgetting turned 
on and off.
The constants used for the experiments were: $\tau = 0.1$,
$\alpha=0.5$, $\beta=0.1$ and a boredom threshold (i.e.,
the value below which a stimuli ceased to be novel)
of 0.4. The parameters of the networks were kept
at the levels found to be optimal in simulations.
The overall qualitative results were the same for all three
networks, although the SOM took longer to produce consistent
output when a new pattern was introduced (owing to the
changes in the spatial pattern in the lag vector) while
the TKM responded to them quickly.

\begin{table*}
\begin{center}
\begin{tabular}{|c|c|c|c|}
\hline
Experiment & Forgetting & Stage & Action \\
\hline
1 & On & Constant On & Robot turns towards it\\
	&& Slow Flashing On & Robot turns towards it\\
	&& Fast Flashing On & Robot turns towards it\\
	&& Constant Off & Robot turns towards it\\
  & Off & Constant On & Robot turns towards it\\
	&& Slow Flashing On & Robot turns towards it\\
	&& Fast Flashing On & Robot turns towards it\\
	&& Constant Off & Robot does not respond\\
\hline
2 & On & Constant On & Robot turns towards it\\
	&& Slow Flashing On & Robot turns towards it\\
	&& Slow Flashing On & If on a different sensor, robot turns towards it\\
  & Off &  Constant On & Robot turns towards it\\
	&& Slow Flashing On & Robot turns towards it\\
	&& Slow Flashing On & If on a different sensor, robot turns towards it\\
\hline
3 & On & Constant On & Robot turns towards it\\
	&& Slow Flashing On & Robot turns towards it\\
	&& Constant On & If on a different sensor, robot turns towards it\\
  & Off &  Constant On & Robot turns towards it\\
	&& Slow Flashing On & Robot turns towards it\\
	&& Constant On & If on a different sensor, robot turns towards it\\
\hline
\end{tabular}
\end{center}
\caption{\textsf{\small{A description of the robots
behaviour in the first series of experiments.}}}
\label{Tab1}
\end{table*}

In table~\ref{Tab1} it can be seen that particular
inputs caused the robot to move even when the
stimulus had been seen before.
This occurred because the stimulus was on a sensor which
had not perceived it previously. This meant that the
robot's attention was changing unnecessarily, so a method to rectify this was devised.
When a stimulus is marked as novel the robot
rotates through $360^\circ$, pausing every
$90^\circ$, so that each of the novelty
filters learns to recognise all the stimuli.
This means that the robot reacts to stimuli
in the same way regardless of which sensor they
impinge on. This functionality can be produced in
other ways, such as using one novelty filter to
monitor all the sensors and adding additional 
memory of what each sensor was seeing to turn the
robot in the appropriate direction.
The output of the network took a few iterations
to stabilise for each new input, and the SOM in
particular occasionally generated spurious
readings, caused by misreading the signals so that
the input vector varied. This was usually because
the sensor polling could not be precisely timed,
so that occasionally the time between readings varied
and so an unexpected input was received.

\subsection{Further Experiments}

In the experiments described previously, all three
clustering networks showed similar qualitative results.
For this reason, further tests were designed to try and
discriminate between the networks. 
The additional experiments performed involved
using flashing lights which flashed at varying
speeds. The neotaxis behaviour of the robot remained
fixed. Two additional patterns of flashing lights were used,
short--short--long--long and short-long-short-long, which the
K{-}Means network and Temporal Kohonen Map both recognised more accurately
than the SOM. The TKM in particular dealt with all
the stimuli very well, but the SOM was occasionally
subject to errors and took longer to respond.
The number of patterns which it is possible for the robot
to learn and recognise is limited by the size of the network.

\section{Conclusions and Future Work \label{Disc}}

The mechanism described here is capable of 
recognising features which vary
in time and habituating to those that are seen
repeatedly. In this way it successfully acts as
a novelty filter, highlighting those stimuli which
are new and directing attention towards them. 
This is a useful ability, since it can 
reduce the amount of data which the robot needs to
process in order to deal with its environment.
However, in the application described here, 
the inputs are fairly clean, the environment being
designed to produce differentiable inputs.

One of the assumptions that is made in this paper is that 
the clustering networks used will reliably separate the
inputs so that new stimuli cause a new neuron to win, and
old stimuli activate the same neuron each time. This is not
necessarily true, and the potential problems this highlights
need to be investigated. Using a growing network such as the
Growing Neural Gas of Fritzke~\cite{Fritzke95} is one solution,
as is using a Mixture of Experts~\cite{Jordan94} in place
of the clustering network, each expert recognising a different
part of the input space. 

In addition, the sensors used here, photocells,
are crude and do not give a great deal of information,
and the robot has very limited memory. To produce a
system which is capable of interacting with real world
environments it will be necessary to use more and better
sensors. 
The next step will be to transfer the system 
onto the Manchester Nomad 200 robot, {\em FortyTwo},
and take advantage of the sensors available, viz. 
sonar, infra-red and a monochrome CCD camera. Before
the novelty filter can deal with this information,
sensor inputs will have to be extensively preprocessed, with
features extracted from the images. Work
using sonar scans taken whilst the robot is exploring
an environment have shown success in applying the
novelty filter to a real world problem (work to be published).

However, once data about the surrounding 
environment can be interpreted, the novelty
filter presented here can be used in an
inspection agent which learns 
a representation of an
environment and can then explore and detect
new or changed features within both that and
similar environments. This is the ultimate 
aim of this research.

\section*{Acknowledgements}
This research is supported by a UK EPSRC Studentship.

\bibliography{thebib,Manbib}
\bibliographystyle{plain}

\end{document}